\documentclass[num-refs]{wiley-article}

\usepackage{siunitx}
\usepackage{amsmath}
\usepackage{romannum}
\usepackage{mathtools}
\usepackage{adjustbox}
\usepackage{graphicx}
\usepackage{xcolor}
\usepackage{dblfloatfix}
\usepackage{amssymb}
\usepackage{algorithm}
\usepackage{algorithmic}
\usepackage{ulem}
\usepackage{multirow}
\usepackage{soul}
\usepackage{threeparttable}

\papertype{Full Paper}

\title{{Q}uantitative {S}usceptibility {M}ap {R}econstruction {U}sing {A}nnihilating {F}ilter-based {L}ow-{R}ank {H}ankel {M}atrix {A}pproach}

\author[1]{Hyun-Seo Ahn}
\author[1\authfn{1}]{Sung-Hong Park}
\author[1\authfn{1}]{Jong Chul Ye}

\affil[1]{Department of Bio and Brain Engineering, Korea Advanced Institute of Science and Technology, Daejeon, South Korea}

\corraddress{Sung-Hong Park PhD,

Department of Bio and Brain Engineering, Korea Advanced Institute of Science and Technology, Rm 1002, CMS (E16) Building, 

291 Daehak-ro, Yuseong-gu, Daejeon, 34141 South Korea
}
\corremail{sunghongpark@kaist.ac.kr}

\fundinginfo{
National Research Foundation of Korea, Grant Number: NRF-2016R1A2B3008104, NRF-2017R1A2B2006526; 

Korea Health Technology R\&D Project, Korea Health Industry Development Institute (KHIDI), funded by the Ministry of Health \& Welfare of South Korea: HI16C1111 
}

\begin{document}

\abbrevs{QSM, quantitative susceptibility mapping; ALOHA, annihilating low-rank Hankel matrix approach; COSMOS, calculation of susceptibility through multiple orientation sampling.
}
\contrib[\authfn{1}]{These authors contributed equally to this work.}

\runningauthor{Hyun-Seo Ahn et al.}

\pagenumbering{arabic}
\maketitle

\begin{abstract}
\textbf{Purpose}: 
Quantitative susceptibility mapping (QSM) inevitably suffers from streaking artifacts caused by zeros on the conical surface of the dipole kernel in k-space. This work proposes a novel and accurate QSM reconstruction method based on k-space low-rank Hankel matrix constraint, avoiding the over-smoothing problem and streaking artifacts.

\noindent\textbf{Theory and Methods}: 
Based on the recent theory of annihilating filter-based low-rank Hankel matrix approach (ALOHA), QSM is formulated as deconvolution under low-rank Hankel matrix constraint in the k-space. The computational complexity and the high memory burden were reduced by successive reconstruction of 2-D planes along three independent axes of the 3-D phase image in Fourier domain. Feasibility of the proposed method was tested on a simulated phantom and human data and were compared with existing QSM reconstruction methods.

\noindent\textbf{Results}: 
The proposed ALOHA-QSM effectively reduced streaking artifacts and accurately estimated susceptibility values in deep gray matter structures, compared to the existing QSM methods.

\noindent\textbf{Conclusion}: 
The suggested ALOHA-QSM algorithm successfully solves the three-dimensional QSM dipole inversion problem by using k-space low rank property with no anatomical constraint. ALOHA-QSM can provide detailed brain structures and accurate susceptibility values with no streaking artifacts.

\end{abstract}

\section{Introduction}

Magnetic susceptibility provides physiological property of tissue different from that of conventional MRI contrasts such as T1w, T2w, and proton density. Magnetic susceptibility is a potential biomarker providing clinically relevant information including tissue iron concentration \cite{B.Bilgic.2012, bilgic2014fast} and venous oxygen saturation level \cite{E.Haacke.2010, fan2014quantitative}. Also, distribution of susceptibility is related to tissue abnormality, which can be used as a diagnosis tool for neuro-degenerative diseases \cite{Acosta.2013} and neuro-surgical planning \cite{E.Winkler.2017}. Phase imaging and susceptibility weighted imaging (SWI) have been used to investigate susceptibility of tissues \cite{E.Haacke.2004}; however, susceptibility signals in these methods are not quantitative and are slightly different from the original susceptibility sources. Quantitative susceptibility mapping (QSM) has been developed to map the original susceptibility sources quantitatively in their exact shapes and locations \cite{L.Rochefort.2008,L.Rochefort.2010}. QSM is widely used since it can be acquired with conventional gradient-echo imaging protocols and offer tissue susceptibility quantitatively with no additional scan.

    From the physical viewpoint, every susceptibility source generates dipole-shaped local phase distribution when placed in a strong magnetic field. Phase images can thus be expressed as the convolution of the susceptibility source and the dipole kernel. This complicated convolutional relationship between phase images and susceptibility images can be simplified as a simple division after the Fourier transform \cite{R.Salomir.2003, J.Marques.2005}. 
    However, in Fourier domain, the dipole kernel generates signal voids near the conical surface that forms an angle of 54.7$^\circ$ along the $B_0$ axis (magic angle), which makes the problem ill-posed. This ill-posed problem can be solved completely by acquiring additional data at three different positions \cite{T.Liu.2009, S.Wharton.2010(1)}; however, it has practical limitations in real clinical application. In many studies, QSM images are acquired along single head orientation by simple replacement of the void k-space region (the conical surface) with tuning parameters (threshold based k-space division; TKD \cite{K.Shmueli.2009,F.Schweser.2013}), homogeneity reconstruction using intrinsic information of gradient echo phase images (homogeneity-enabled incremental dipole inversion; HEIDI \cite{F.Schweser.2012}), reconstruction based on sparsity of the estimated susceptibility maps (compressed sensing compensation; CSC \cite{B.Wu.2012}), or regularization using structural edge information of magnitude images (morphology enabled dipole inversion; MEDI \cite{J.Liu.2012,T.Liu.2012}). However, current QSM methods still suffer from streaking artifacts, over-smoothing, and/or errors induced by magnitude images \cite{Y.Wang.2015}.

    Recently, low-rank Hankel matrix approaches have been extensively studied for reconstruction of sparse MR images \cite{Shin.2014, Haldar.2014, Haldar.2016, K.Jin.2016, K.Jin.2017, K.Jin.2015, Kim.2017, D.Lee.2016, J.Lee.2016, Mani.2017, Ongie.2017(1),Ongie.2016, Ongie.2017(2)}, with proven theoretical optimality \cite{J.Ye.2017}. Conventional compressed sensing methods reconstruct images from down-sampled data by exploiting sparsity in a specific transform domain including wavelet transform and total variation (TV) \cite{M.Lustig.2008}. Low-rank Hankel matrix approaches are also based on sparsity of the data; however, instead of using sparsity as a regularization term, these approaches use the low-rankness of the weighted Hankel matrix \cite{K.Jin.2016, J.Ye.2017}. Low-rank Hankel matrix approaches have been categorized into two classes: a penalized least squares approach using low-rank Hankel matrix constraint \cite{Shin.2014, Haldar.2014, Kim.2017, Mani.2017, Ongie.2017(1), Ongie.2016, Ongie.2017(2)} and the direct k-space interpolation method using annihilating filter-based low-rank Hankel matrix (ALOHA) \cite{K.Jin.2016, K.Jin.2017, K.Jin.2015, D.Lee.2016, J.Lee.2016}. For example, ALOHA  has many applications such as acceleration of MR image acquisition \cite{K.Jin.2016}, EPI ghost correction \cite{J.Lee.2016}, MR parameter mapping \cite{D.Lee.2016}, MRI artifact correction \cite{K.Jin.2017}, and image inpainting \cite{K.Jin.2015}. 
    However, these structured low-rank Hankel matrix approaches have never been applied to solving the dipole inversion problem of QSM. 

    In this study, we propose a new k-space based 3-D QSM reconstruction method, referred to as ALOHA-QSM, for high accuracy and minimal artifacts. ALOHA-QSM has been evaluated on a numerical phantom and human brain data in public domains as well as  additional human brain data acquired from five subjects, in comparison with other existing QSM methods.

\section{Theory}

\subsection{Dipole inversion}

 When a substance is placed in a magnetic field, it tends to repel or assist the surrounding magnetic field lines, a phenomenon called magnetic susceptibility ($\chi$). Specifically, paramagnetism ($\chi>0$) and diamagnetism ($\chi<0$) respectively attract and repel surrounding magnetic field lines. In a gradient echo sequence, this field distribution induces evolution of MR phase signals during the echo time in a spatially different manner, resulting in a dipole-shaped phase distribution around the susceptibility source.
    Hence, the relationship between the susceptibility source and its resultant phase images can be expressed using convolution as follows:
\begin{equation}\label{eq:relation}
{\theta(\vec{r})}= (2{\pi}{\bar{\gamma}}\cdot H_{0}\cdot{TE})\cdot({D(\vec{r})}*{\chi(\vec{r})}), 
\end{equation}
    where $\theta(\vec{r})$ is the phase signal generated by the susceptibility source, $\bar{\gamma}$ is the gyromagnetic ratio, $H_0$ is strength of the applied magnetic field, TE is an echo time of the gradient echo,  $*$ is a convolution operator,  and $D(\vec{r})$ is the dipole kernel which represents local magnetic field shift induced by the susceptibility source, given by
\begin{equation}\label{eq:Green}
D(\vec{r})=\frac{1}{4\pi}\frac{3\cos^{2}\Theta-1}{r^3},
\end{equation}
    where $\Theta$ is the angle between $\vec{r}$ and the direction of the main magnetic field.

    Based on  Eq.(\ref{eq:relation}), magnetic susceptibilities can be estimated as deconvolution of the phase image with the dipole kernel. However, applying deconvolution on Eq.(\ref{eq:relation}) directly is computationally too demanding, because all of these calculations are based on 3-D. Instead, one can change this problem into the simple division problem on the Fourier domain. 
    For simplicity, we will denote Fourier domain of phase images and susceptibility images as "k-space" and their axes as $k_x-k_y-k_z$ in k-space and $x-y-z$ in the image domain.
    Then, the dipole kernel in Eq.(\ref{eq:Green}) can be transformed into the simple k-space filter as follows:
\begin{equation}\label{eq:dipole1}
    \hat{D}(\vec{k})=\frac{1}{3}-\frac{k_z^2}{\vec{k}^2}
\end{equation}
    where $\vec{k}$ is a k-space vector $\vec{k}=[k_x, k_y, k_z]^T$, and $\vec{k}^2=k_x^2+k_y^2+k_z^2$.
    Also, by denoting $\varphi=\theta/(2{\pi}{\bar{\gamma}}\cdot H_{0}\cdot{TE})$, the relationship between the phase and the magnetic susceptibility in Eq.(\ref{eq:relation}) can be simplified into:
\begin{equation}\label{eq:dipole2}
    \hat{\chi}(\vec{k})=\frac{\hat{\varphi}(\vec{k})}{\hat{D}(\vec{k})},
\end{equation}
    where $\hat{\chi}(\vec{k})$ and $\hat{\varphi}(\vec{k})$ are the k-spaces of the susceptibility image and phase image, respectively. However, $\hat{D}(\vec{k})$ becomes zero on the conical surface: $\vec{k}=0$ or $k_z^2 =1/3 \vec{k}^2$, on which the direct inversion using Eq.(\ref{eq:dipole2}) is not valid. 
	Many studies have been conducted to solve this inversion problem to calculate the optimal susceptibility image satisfying Eq.(\ref{eq:dipole2}). This is often called the dipole inversion problem. 
    For example, the calculation of susceptibility through multiple orientation sampling (COSMOS) is the most representative and obvious method to solve the problem \cite{T.Liu.2009}.  
    It oversamples data with multiple head orientations and eliminate all the null points at the conical surface.
    COSMOS can serve as the ideal solution for the dipole inversion problem. However, it requires patients to be scanned at least three times with different head orientations. From this reason, it is not easy to apply COSMOS clinically. 

    There are many approaches to address this problem with the data acquired at a single head orientation.
    Truncated k-space division (TKD) is a simple and intuitive method \cite{K.Shmueli.2009,F.Schweser.2013}. Some portions of the dipole kernel filter near the conical surface are replaced with the simple threshold value ($a$) as follows: 
    \begin{equation}\label{eq:TKD}
    \begin{aligned}
    \hat{D}(\vec{k},a) = \begin{cases}
    \hat{D}(\vec{k})  \quad &\text{when}\,\,|\hat{D}(\vec{k})|>a \\
    a \cdot\text{sign}(\hat{D}(\vec{k})) \quad &\text{when}\,\,|\hat{D}(\vec{k})| \leq a\\
    \end{cases}
    \end{aligned}
\end{equation}
    However, it causes discontinuity and vanishing surfaces on the resultant k-space, which leads to streaking artifacts in the corresponding image domain. Also, susceptibility values are often underestimated when a high threshold value is used \cite{F.Schweser.2013}.

    Alternatively, additional information from the data can be exploited to handle this uncertainty near the conical surface. One of the popular and powerful way is to use magnitude images. Morphology enabled dipole inversion (MEDI) assumes that structures of susceptibility sources are well reflected on the magnitude information \cite{T.Liu.2011,J.Liu.2012}. 
	Specifically, the following equation is derived to reconstruct the optimal susceptibility map by assuming that the structural boundaries on susceptibility images coincide with those on the magnitude images.
\begin{equation}\label{eq:MEDI}
     \min_{\chi}{\parallel W(\varphi-{FT^{-1}}(\hat{D}\cdot FT({\chi})))\parallel^2 + \lambda\parallel M\bigtriangledown\chi\parallel_1}, 
\end{equation}
    where \noindent$W$ is the weighting mask derived from the magnitude image, $FT$ is the Fourier-transform of the matrix, $FT^{-1}(\cdot)$ is the inverse Fourier-transform of the matrix, $\lambda$ is the regularization parameter, $\bigtriangledown$ is a 3-D gradient operator, and $M$ is a binary mask of structural boundaries on the magnitude image.
      Streaking artifacts in susceptibility images can be also minimized by subtracting streaking artifacts iteratively (iLSQR) \cite{W.Li.2011,W.Li.2015}. This method assumes that the streaking artifacts are originated from inaccuracies of inversion at the truncated k-space regions. By iteratively subtracting this artifact-originated image, artifact-reduced susceptibility image is obtained.

    The aforementioned algorithms are reported to estimate susceptibility values accurately \cite{Y.Wang.2015}. However, they still have many limitations. K-space based methods such as TKD or iLSQR show remaining streaking artifacts in images. MEDI may induce some bias originated from the difference between the magnitude and susceptibility images. 

\subsection{Deconvolution for QSM using ALOHA}
    
    Compressed sensing (CS) is one of the most powerful signal reconstruction methods that can recover accurate MR images from down-sampled k-space measurements. 
	Important requirements for applying CS algorithm to MR image reconstruction are sparsity of the data in some transform domain and incoherence of the sensing matrix \cite{M.Lustig.2008}.

    Inspired from the CS theory, low-rank Hankel matrix approaches have been proposed to complete the acquired incomplete data matrix by using its low-rankness \cite{Shin.2014, Haldar.2014, Haldar.2016, K.Jin.2016, K.Jin.2017, K.Jin.2015, Kim.2017, D.Lee.2016, J.Lee.2016, Mani.2017, Ongie.2017(1),Ongie.2016, Ongie.2017(2), J.Ye.2017}, which is directly related to the sparsity in the image domain \cite{J.Ye.2017, Ongie.2017(1), Ongie.2016}.
   This implies that an artifact-free image can be recovered from the incomplete or distorted k-space measurements by constructing Hankel structured matrix and exploiting its low-rankness.
   However, the dipole inversion problem in QSM is quite different from the conventional CS problem: the under-determined region of the dipole kernel filter ($\hat{D}$) clearly forms a conical shape.
    Fortunately, this region occupies only a small portion (about 10\%) of the whole k-space compared to the down-sampled region of the conventional CS acceleration ($\geq$50\%). In previous studies, CS successfully reconstructed the susceptibility images from the phase images \cite{B.Wu.2012,C.Langkammer.2017}.   
    Accordingly, we tried to apply the low-rank Hankel matrix approach, the ALOHA algorithm, for solving the QSM dipole inversion problem. 
    
    Specifically, the weighted k-space of artifact-suppressed susceptibility images can be estimated by formulating an equation as follows:
    \begin{equation} \label{eq:ALOHA_QSM}
\min_{\hat{\chi}_w} {\parallel \hat{\varphi}_w - \hat{D} \odot \hat{\chi}_w \parallel}^2 + \lambda \textrm{RANK} (\mathcal{H}(\hat{\chi}_w)), \\
\end{equation}
    where $\hat{\varphi}_w$ is the weighted k-space of the measured phase image, $\hat{\chi}_w$ is the weighted k-space of the estimated susceptibility image, $\lambda$ is the regularization parameter,
    $\odot$ is the element-wise multiplication operator,
     $\mathcal{H}$($\hat{\chi}_w$) denotes a Hankel structured matrix of the weighted susceptibility k-space, and RANK($\cdot$) refers to the matrix rank. The first term of Eq.(\ref{eq:ALOHA_QSM}) is the data fidelity term, and the second term is the additional regularization term based on the low-rankness of the Hankel structured matrix. Note that the problem is formulated entirely in the Fourier domain, and the weighted k-space comes from Haar wavelet spectrum. 
	 Specifically, the Hankel structured matrix of an \textit{n-}dimensional vector $x$ has the following structure:
	 \begin{equation}\label{eq:Hankel}
	 \mathcal{H}(x)=\begin{pmatrix}
	 x[0] & x[1] & \cdots & x[n] \\
	 x[1] & x[2] & \cdots & x[0] \\
	 \vdots & \vdots & \ddots & \vdots \\
	 x[n] & x[0] & \cdots & x[n-1]
	 \end{pmatrix}
	 \end{equation}

    Since the rank term is not convex, it is difficult to estimate the low-rank matrix by solving Eq.(\ref{eq:ALOHA_QSM}) directly.
	Hence, as suggested in \cite{K.Jin.2016,J.Ye.2017}, the matrix factorization approach \cite{Signoretto.2013} is used for estimating $\textrm{RANK}(\mathcal{H}(\hat{\chi}_w))$ based on the following observation \cite{Srebro.2004}:
    \begin{equation}\label{eq:Nuclear}
    \parallel{A}\parallel_{*}=\min_{U,V:A=UV^H}{\parallel{U}\parallel^2_F+\parallel{V}\parallel^2_F},
    \end{equation}
    where ($\cdot$)$^H$ is the Hermitian transpose of a matrix. By adopting this idea, Eq.(\ref{eq:ALOHA_QSM}) can be solved easily as shown below, although still non-convex. 
\begin{equation} \label{eq:ADMM}
    \min_{\hat \chi_w, U,V:\mathcal{H}( \hat{\chi}_w)=UV^H}{\frac{1}{2} {\parallel \hat{\varphi}_w-\hat{D} \odot \hat{\chi}_w \parallel}^2 + \frac{\lambda}{2} ({\parallel U \parallel}_F^2 + {\parallel V \parallel}_F^2 ) }.
\end{equation}
    There are three variables in this equation: $\hat\chi_w$, $U$, and $V$. Hence, we employed the alternating direction method of multiplier (ADMM) algorithm with the following Lagrangian \cite{Boyd.2011}:
\begin{equation} \label{eq:Lagran}
\begin{split}
    L(U,V,\hat{\chi}_w,\Lambda) = 
    \frac{1}{2} {\parallel \hat{\varphi}_w-\hat{D} \odot \hat{\chi}_w \parallel}^2 + \frac{\lambda}{2} ({\parallel{U}\parallel}_F^2 
    + {\parallel{V}\parallel}_F^2 ) 
    \\ + \frac{\mu}{2} {\parallel \mathcal{H}(\hat{\chi}_w) - UV^H +\Lambda \parallel^2},
\end{split}
\end{equation}
    where $\mu$ is the hyper-parameter of ADMM and $\Lambda$ is the Lagrangian to be updated. 
    Variables of Eq.(\ref{eq:Lagran}) are updated by
\begin{equation}
\label{eq:ADMM_detailed}
\begin{split}
\hat{\chi}_w ^ {(n+1)} &=\frac{\hat{D}^* \odot \hat{\varphi}_w + \mu (\mathcal{H}^{*} (U^{(n)} V^{(n)H} - \Lambda^{(n)})}
    {{ |\hat{D}| }^2 + \mu} \\
U^{(n+1)} &= \mu (\mathcal{H}(\hat\chi_w^{(n+1)})+\Lambda^{(n)}) V^{(n)} (\lambda I +\mu V^{(n)H} V^{(n)})^{-1}\\
V^{(n+1)} &= \mu(\mathcal{H}(\hat\chi_w^{(n+1)})+\Lambda^{(n)})^{H} U^{(n+1)} (\lambda I + \mu U^{(n+1)H} U^{(n+1)})^{-1}\\
\end{split}
\end{equation}
and the Lagrangian update is given by
\begin{equation}
    \label{eq:Lagrangian}
    \Lambda^{(n+1)} = \mathcal{H}({\hat{\chi}_w ^{(n+1)}}) - U^{(n+1)} V^{(n+1)H} + \Lambda^{(n)}.\\
\end{equation}

The dipole inversion requires storage of 3-D Hankel matrix in memory and factorization of its matrix, which is computationally demanding. While circulant approximation of Hankel matrix is recently proposed to handle this issue \cite{Ongie.2017(2)}, we propose to overcome this limitation by applying the 2-D ALOHA algorithm to the 3-D data successively along  three different directions ($k_x, k_y, k_z$). 
    
    Applying 2-D inverse Fourier transform to the 3-D k-space along $k_z$ direction will generate a 3-D data in the domain of $x-y-k_z$, where the central and peripheral slices contain low-frequency and high-frequency information, respectively, along the axial direction. This image can be also sparsified using Haar wavelet transform, indicating the existence of the corresponding annihilating filter, because the Hankel matrix derived from the weighted k-space along a specific direction shows low-rank property \cite{J.Ye.2017}. Hence, the 3-D k-space can be split up into multiple 2-D planes and each 2-D plane is processed using the 2-D version of ALOHA-QSM. The same trick can be applied along the other two directions as well, i.e., $k_y$ or $k_x$ direction. To minimize variations according to the choice of the processing direction, this 2-D processing was applied once along each of the three directions successively in this study. No further iteration was used since it converged after the first iteration (Supporting Information Fig.S1). This procedure can be understood as successive relaxation methods as used in iterative coordinate descent method (ICD) \cite{bouman1996unified} and space alternating generalized expectation maximization (SAGE) algorithm \cite{fessler1994space}, whose convergence property has been proven.

\subsection{Overview of the ALOHA-QSM method}

\begin{figure*}[ht]
\includegraphics[width=\textwidth]{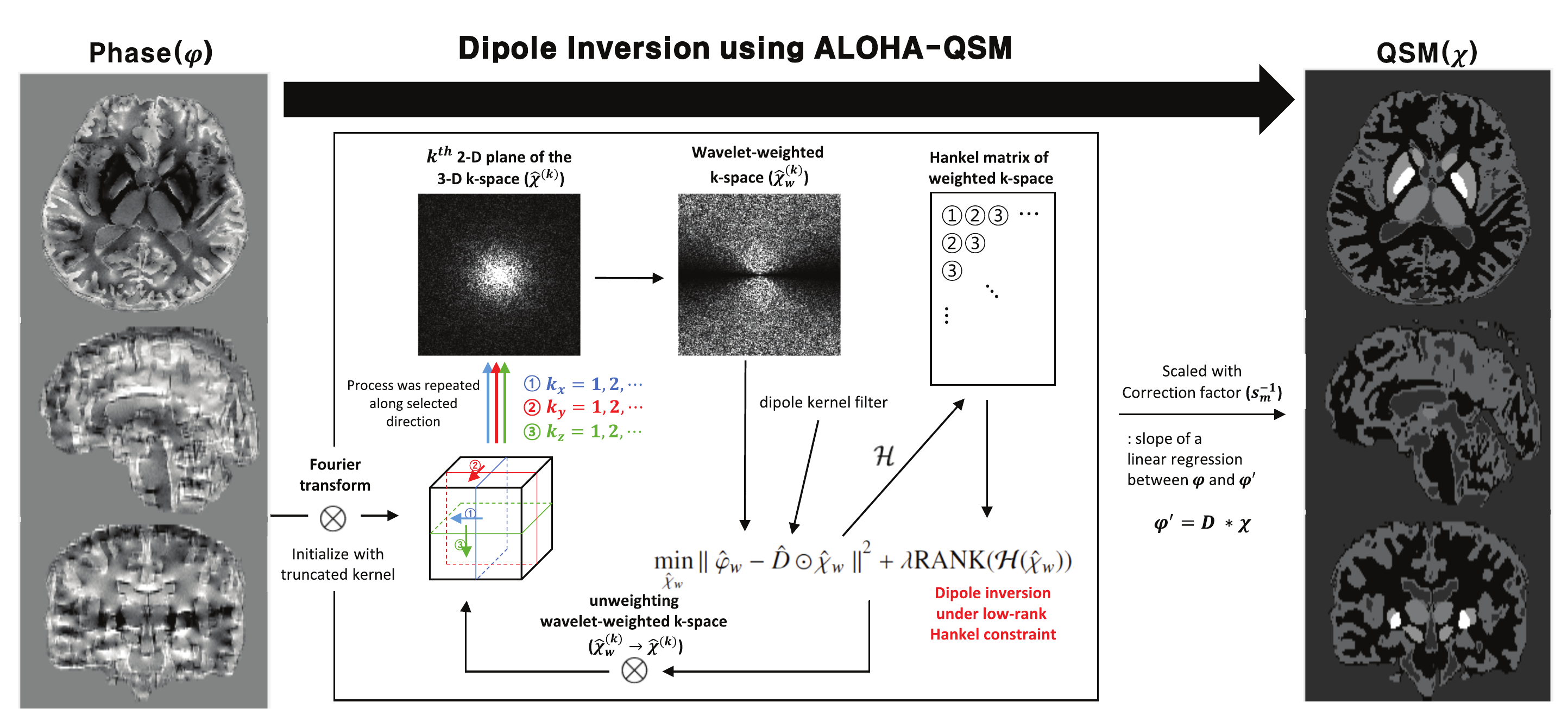}
\caption{The overall flow diagram for the ALOHA-QSM algorithm on the dipole inversion problem. Phase image was Fourier transformed into 3-D k-space. Then, the Haar wavelet spectrum was used for weighting the slice of k-space. The problem was solved using ADMM algorithms, and Haar wavelet spectrum was unweighted. Algorithm was applied to every slices following selected direction, then successively implemented along the other two remaining directions. For the first loop, truncated dipole kernel was used for data initialization and for next loops, results from the previous loop was used. Finally, the correction factor was multiplied to compensate the systematic underestimation.
}\label{fig:scheme}
\end{figure*}

Detailed process of ALOHA-QSM method is described in Fig.\ref{fig:scheme} and Appendix (Algorithm \ref{algorithm:ALOHAQSM}).
Note that the 3-D k-space of phase image is used as input for the algorithm. 
At first, the 3-D k-space was initialized with the TKD method, where the threshold value was set to 0.1. Next, one of the k-space directions ($k_x,k_y,k_z$) was selected, and each 2-D plane of the 3-D k-space along the selected direction was processed by 2-D ALOHA. For each 2-D plane, Haar wavelet weighting was applied to exploit the sparsity of the selected plane \cite{K.Jin.2016}. 
Then Eq.(\ref{eq:ALOHA_QSM}) was formulated and then an artifact-suppressed Haar wavelet weighted k-space plane was reconstructed by using ADMM algorithm in Eq.(\ref{eq:ADMM_detailed}) and Eq.(\ref{eq:Lagrangian}). After that, Haar wavelet weighting was removed.
This process was applied to all the planes along the first-selected direction individually. The same whole procedure was repeated for the other two k-space directions successively, providing a new artifact-suppressed 3-D k-space of the susceptibility image.

However, due to the null points at the conical surface of the dipole kernel's k-space, the estimated susceptibility values were underestimated. To address this problem, we employed a correction factor $s_m$ by calculating the slope of a least-squares regression between the original phase image and the phase image obtained by applying the forward model \cite{J.Marques.2005} on the susceptibility image reconstructed with the ALOHA-QSM method. The final QSM image was generated after compensation of the intensity scale with the correction factor.

\section{Methods}

\subsection{Public data from repository}
	To compare the performance of ALOHA-QSM with other methods, the numerical phantom and the \textit{in vivo} human brain data were downloaded from the repository (http://weill.cornell.edu/mri/pages/qsmreview.html and http://qsm2016.com/) \cite{Y.Wang.2015,C.Langkammer.2017}. 
    For the phantom and \textit{in vivo} data, the resolution was 0.9375 $\times$ 0.9375 $\times$ 1.5 mm$^3$ and 1.06 $\times$ 1.06 $\times$ 1.06 mm$^3$, and the matrix size was 256 $\times$ 256 $\times$ 98 and 160 $\times$ 160 $\times$ 160, respectively. 
	For the numerical phantom data, noise with SNR=10 was added to the images to simulate a more realistic situation.
	For the \textit{in vivo} human brain, datasets acquired with COSMOS along 12 different head orientations were additionally downloaded and used as comparison reference \cite{C.Langkammer.2017}. 
    
\subsection{Experimental data for \textit{in vivo} human brain}
	In order to conduct further detailed quantitative comparison between ALOHA-QSM and other QSM methods, \textit{in vivo} human brain images were obtained using a 12-channel head coil on a 3T MRI scanner (MAGNETOM Verio, Siemens Healthcare, Erlangen, Germany). Five healthy volunteers (5 males, 21-25 years old) were scanned with conventional multi-echo GRE sequences at a single head orientation. Sequence parameters were: bandwidth $=150$ Hz/px for all echoes, TR$=43$ ms, TE$_1=9.35$ ms, $\Delta$TE$=8.94$ ms, number of echoes = 3, matrix size $=320 \times 240 \times 128$, resolution $=0.75 \times 0.75 \times 1.1 mm^3$, 
	flow compensation along all the three directions, slice oversampling $=25\%$, and flip angle = $14^\circ$. In our study, partial Fourier along phase and slice encoding directions with factor $=6/8$ was used, which did not cause significant changes on phase images (Supporting Information Fig.S2). Total acquisition time was $15$ min $30$ sec for each subject.
 
\subsection{Data Processing}
	For human brain data acquired from multi-coils, coil combination was executed first. All the data were processed as follows. Brain mask was extracted from magnitude images, using brain extraction tool (BET) of FSL \cite{S.Smith.2002}. The phase images from multi-coils were combined by Hermitian inner product \cite{MA.Bernstein.1994}. Multi-echo phase images were corrected with non-linear frequency map estimation \cite{T.Liu.2013,B.Kressler.2010,L.Rochefort.2008}. Laplacian-based phase unwrapping was employed to avoid abrupt phase jumps\cite{W.Li.2011,Schofield.2003}. Varied version of Sophisticated Harmonic Artifact Reduction on Phase data (V-SHARP) was applied for eliminating background phase variations \cite{W.Li.2011,B.Wu.2012,W.Li.2014}.

    GeForce GTX 1080 graphic card was used as graphic processor unit (GPU) and i7-6700k was used as central processing unit (CPU). All the codes were written in MATLAB 2017b (Mathwork, Natick). Computed Unified Device Architecture (CUDA) was employed for GPU to accelerate the algorithm.

	Discrepancy principle was used to determine optimal hyper-parameters for ALOHA-QSM reconstruction \cite{morozov1966solution,bauer2011comparingparameter}. 
	Parameter $\mu$ was investigated from $10^{-3}$ to $10^{0}$ with multiplicative step size of $10^{0.2}$.
	For each $\mu$, $\lambda$ was increased from $10^0$ to $10^4$ with multiplicative step size of $10^{0.2}$. Root-mean squared error (RMSE) between the acquired phase image ($\varphi$) and the reconstructed phase image with ALOHA-QSM ($\varphi'$) was calculated, and the point where RMSE equals the standard deviation in the CSF of phase image for the first time was selected (phantom: $\lambda=10^{1.4}$, $\mu=10^{-1.8}$; human data from repository: $\lambda=10^{1.4}$, $\mu=10^{-2.2}$; human data from experiment: $\lambda=10^{2.4}$, $\mu=10^{-2.2}$). RMSE was calculated as
	\begin{equation}
	    \sqrt{\frac{\sum_{i=1}^n{(x_{r,i}-x_{t,i})^2}}{\sum_{i=1}^n{(x_{t,i})}}},
	\end{equation}
	where $x_{t,i}$ and $x_{r,i}$ represent pixel intensities of the reference image and the image to be calculated, and $n$ is the total number of pixels in the image.
	
	The performance of the ALOHA-QSM method was evaluated in comparison with three representative QSM reconstruction methods: TKD \cite{K.Shmueli.2009}, iLSQR \cite{W.Li.2011,W.Li.2015}, and MEDI \cite{T.Liu.2011,J.Liu.2012}.  
	Reconstruction parameters were as follows: $a=0.1$ for TKD, 30 iteration steps for iLSQR, and for MEDI: $\lambda=600$ (numerical phantom and experimental huamn brain) and $\lambda=6000$ (human brain data from repository) \cite{W.Li.2015,Y.Wang.2015}.

    Mean susceptibility values were quantified in the regions of deep brain gray matter structures: substantia nigra, red nucleus, globus pallidus, putamen, and head of caudate nucleus. ROIs of the gray matter structures were manually segmented from a slice of brain susceptibility images. The mean susceptibility values of TKD, iLSQR, MEDI, and ALOHA-QSM in the structures were compared each other for data from the five \textit{in vivo} human brains. Also, the susceptibility values from ALOHA-QSM was compared with those from literature \cite{B.Bilgic.2012,Lim.2013,T.Liu.2011, S.Wharton.2010(1)}. 
    For the data from repository, linear regression was used to indicate how well the QSM methods estimated susceptibility as reference to the ground truth provided in the repository, and RMSE was calculated for the quantitative evaluation.

\section{Results}

\subsection{Parameter determination}
    
\begin{figure*}[!ht]
\centering
\includegraphics[width=\textwidth]{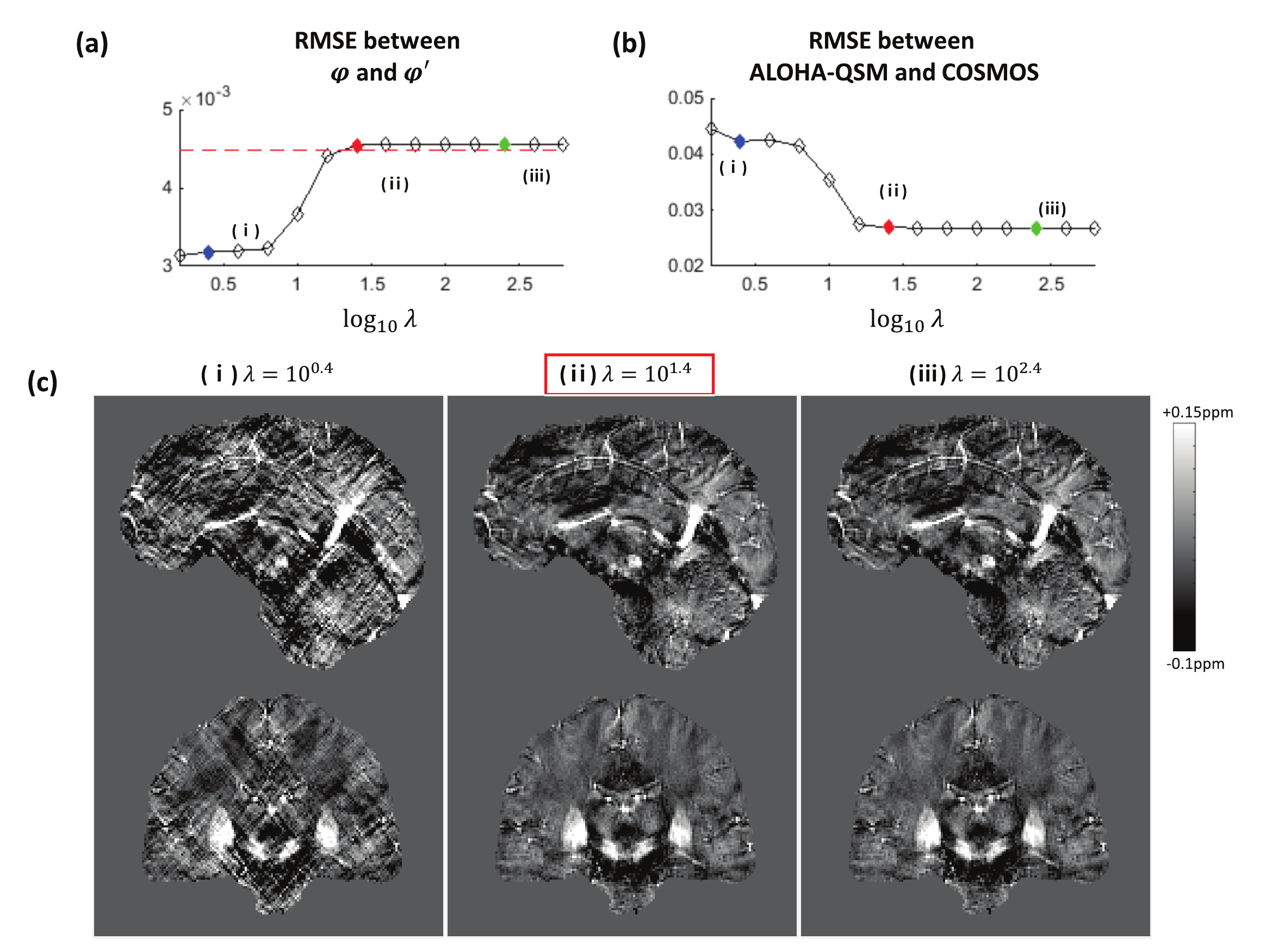}
\caption{
Parameter for ALOHA-QSM was optimized. (a) Root-mean squared error (RMSE) of the acquired phase image ($\varphi$) and the reconstructed phase image ($\varphi’$). Red line is the measured noise level in the CSF from phase image. (b) RMSE between susceptibility maps from ALOHA-QSM and COSMOS was compared for each $\lambda$. 
(c) At three points ($\lambda=10^{0.4},10^{1.4},10^{2.4}$), their sagittal section of the QSM map (top) and the k-space (bottom) were displayed.
}\label{fig:acq_paramopt}
\end{figure*}

	From Eq.(\ref{eq:Lagran}), we can find that the $\lambda$ controls weighting of the matrix factorization of the rank estimation, and the $\mu$ determines iterative updates on the ADMM algorithm. 
	For each $\mu$, RMSE between the acquired phase image and the reconstructed phase image increased at first and converged as $\lambda$ increases. 
	RMSE coincided with the noise level in the phase image at $\mu=10^{-2.2}$ for the \textit{in vivo} human data from repository (Fig.\ref{fig:acq_paramopt}a). With small $\lambda$, streaking artifacts increased in susceptibility maps because of the increased proportion of the data fidelity term in Eq.(\ref{eq:ADMM}), and the result became similar to that of TKD. 
	With $\lambda$ larger than the optimal point, RMSE converged and susceptibility maps showed no discernible changes (Fig.\ref{fig:acq_paramopt}b,c).

\subsection{Comparison of QSM reconstruction methods}

\begin{figure*}[!ht]
\centering
\includegraphics[height=\textwidth]{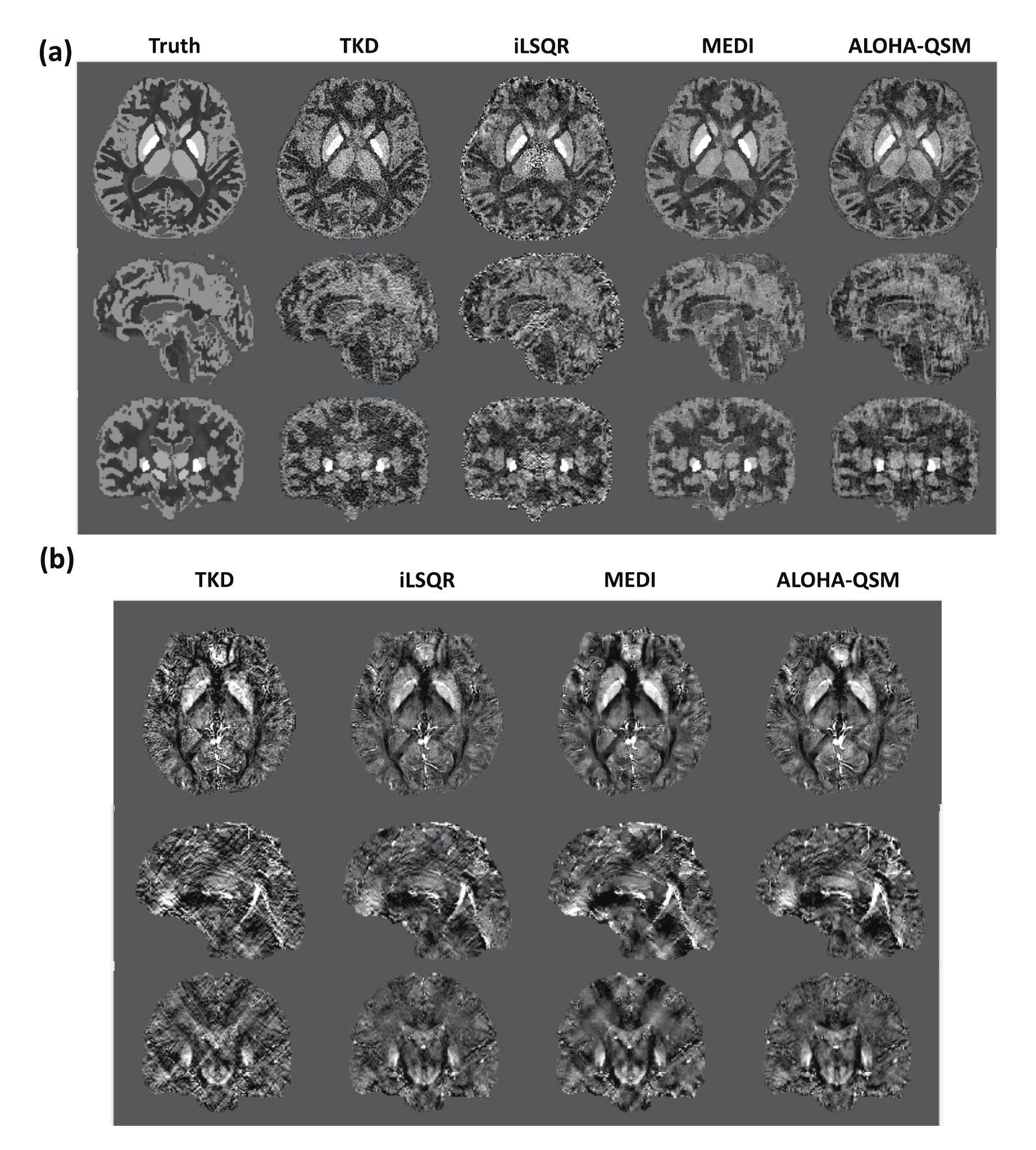}
\caption{Susceptibility maps generated by True, TKD, iLSQR, MEDI, and ALOHA-QSM were compared from (a) the numerical phantom and (b) the \textit{in vivo} human brain.
}\label{fig:comparison}
\end{figure*}

\begin{figure*}[!ht]
\centering
\includegraphics[width=\textwidth]{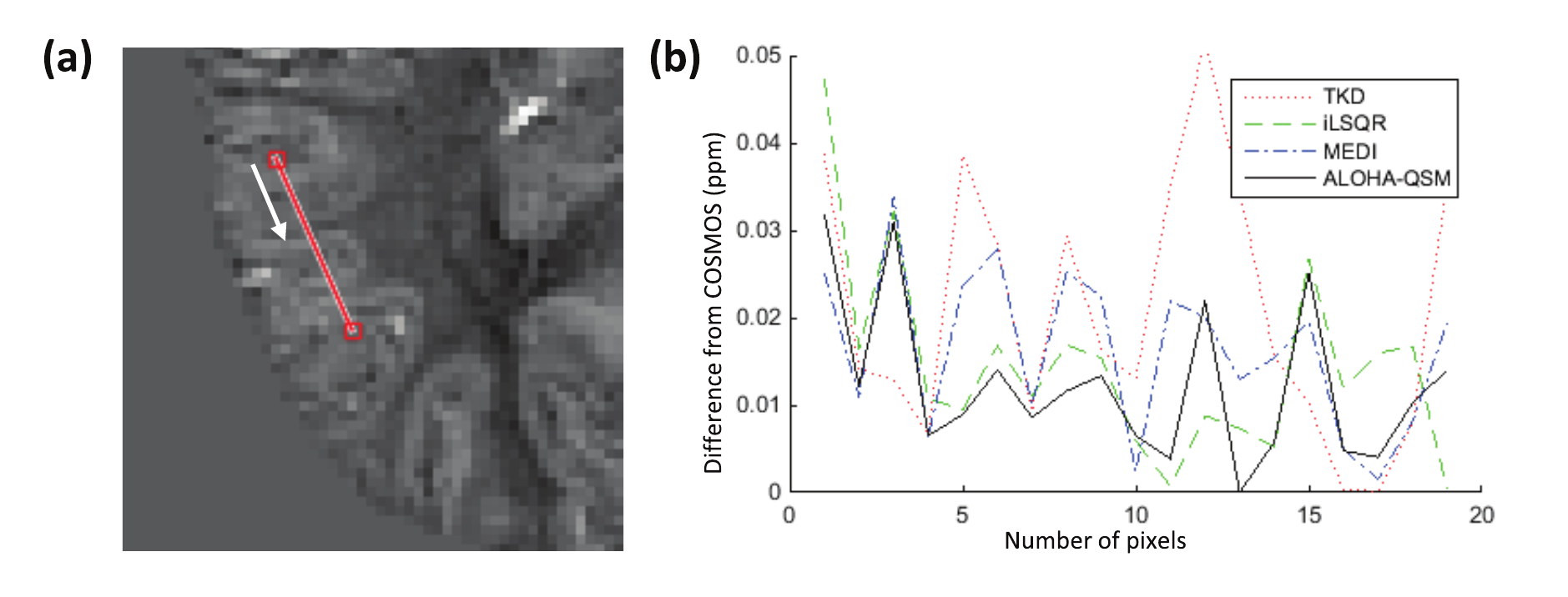}
\caption{(a) Enlarged axial slice image in brain cortex. (b) The susceptibility difference referenced to COSMOS along the red line in (a).
}\label{fig:profile}
\end{figure*}

	Performance of TKD, iLSQR, MEDI, and ALOHA-QSM was compared for the numerical phantom and the human brain data (Fig.\ref{fig:comparison}, Table.\ref{table:comparison}).
	The susceptibility images of TKD and iLSQR were noisier than those of MEDI or ALOHA-QSM, with apparent streaking artifacts (Fig.\ref{fig:comparison}a).  
	In Fig.\ref{fig:comparison}b, reconstructed susceptibility maps of \textit{in vivo} human brain were displayed. TKD showed severe streaking artifacts and obscured structural details. iLSQR reconstructed susceptibility images well, but detailed structures of veins and gray matter were not clearly shown.
	For MEDI, streaking artifacts were better eliminated than TKD and iLSQR, and vascular structures were clearly shown in the sagittal section. However, MEDI showed digitization artifacts.
	ALOHA-QSM reliably reconstructed susceptibility image with clear brain structures and streaking artifacts were removed effectively.
	Fig.\ref{fig:profile}b represents the susceptibility difference from COSMOS along the red line in the brain cortex shown in Fig.\ref{fig:profile}a. iLSQR and ALOHA-QSM matched the reference (COSMOS) well, while the profile of TKD was quite different from that of the reference.

	For quantitative comparison, linear regression coefficients ($R^2$, slope) and RMSE were compared between the dipole inversion methods for the simulated brain and \textit{in vivo} brain (Table.\ref{table:comparison}). The Pearson correlation coefficient ($R^2$) of the \textit{in vivo} brain was lower than that of the simulated brain, presumably due to the reference (COSMOS) acquired from multiple head orientations for the \textit{in vivo} brain.
	For the simulated brain, MEDI showed the largest $R^2$ and the smallest RMSE. For the \textit{in vivo} brain, ALOHA-QSM showed the largest $R^2$ value and the smallest RMSE. After application of the intensity correction, ALOHA-QSM showed more reasonable regression slope (0.89) but slightly increased RMSE, which was still lower than the other methods and was not concentrated in a specific brain region or structure.

\begin{figure*}[ht]
\centering
\includegraphics[width=\textwidth]{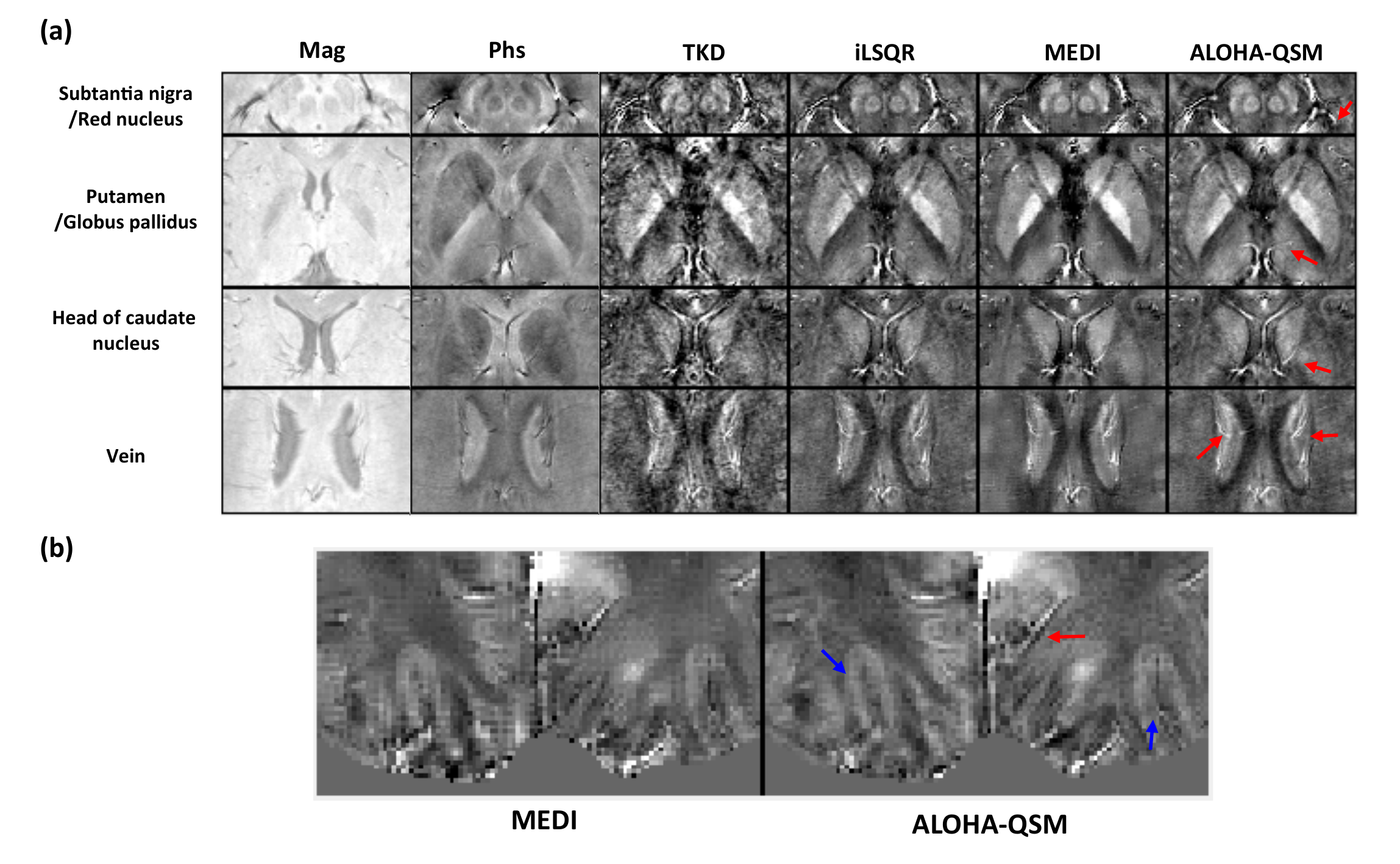}
\caption{
(a) Detailed brain structures with magnitude image, phase image, and susceptibility image for each QSM reconstruction method. (b) Enlarged cortex image in MEDI and ALOHA-QSM. Red arrows indicate small complex vascular structures and blue arrows indicate complicated folded structures.
}\label{fig:invivoroi}
\end{figure*}

\begin{figure*}[!ht]
\centering
\includegraphics[width=\textwidth]{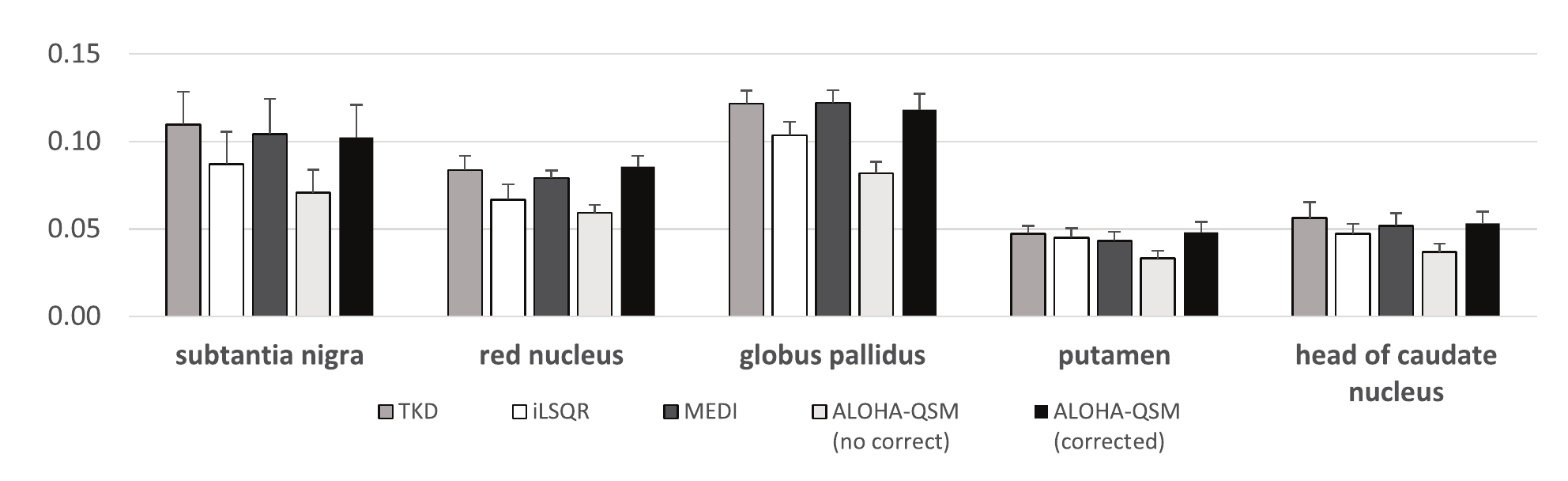}
\caption{
Estimated susceptibilities in deep gray matter structures from five healthy volunteers were compared for each QSM reconstruction method. 
}\label{fig:quantcomp}
\end{figure*}

	For further detailed comparison on reconstructed susceptibility images, some distinctive brain structures are displayed in Fig.\ref{fig:invivoroi}. ALOHA-QSM showed clear descriptions on structural details compared to TKD and iLSQR, especially for deep gray matter structures. MEDI also conserved gray matter structures well, however, images of MEDI were noisier with digitization artifacts than those of ALOHA-QSM, which was clearly shown in the enlarged images (Fig.\ref{fig:invivoroi}b). In addition, complicated folded structures (blue arrow) and small complex vascular structures (red arrow) were more successfully expressed in ALOHA-QSM.
	For deep gray matter structures, ALOHA-QSM provided susceptibility values similar to those of the other QSM methods by using a correction factor (Fig.\ref{fig:quantcomp}). Although there were variations in the susceptibility values for each QSM reconstruction method in literature, ALOHA-QSM showed a reasonable range of quantitative estimates (Table.\ref{table:roi}).

\section{Discussion}
    In this study, we proposed a novel QSM reconstruction algorithm that employs the k-space low-rank Hankel matrix constraint. The proposed algorithm successfully reconstructed susceptibility images for the numerical phantom and human brains acquired at a single head orientation.

\begin{figure*}[hb]
\centering
\includegraphics[width=\textwidth]{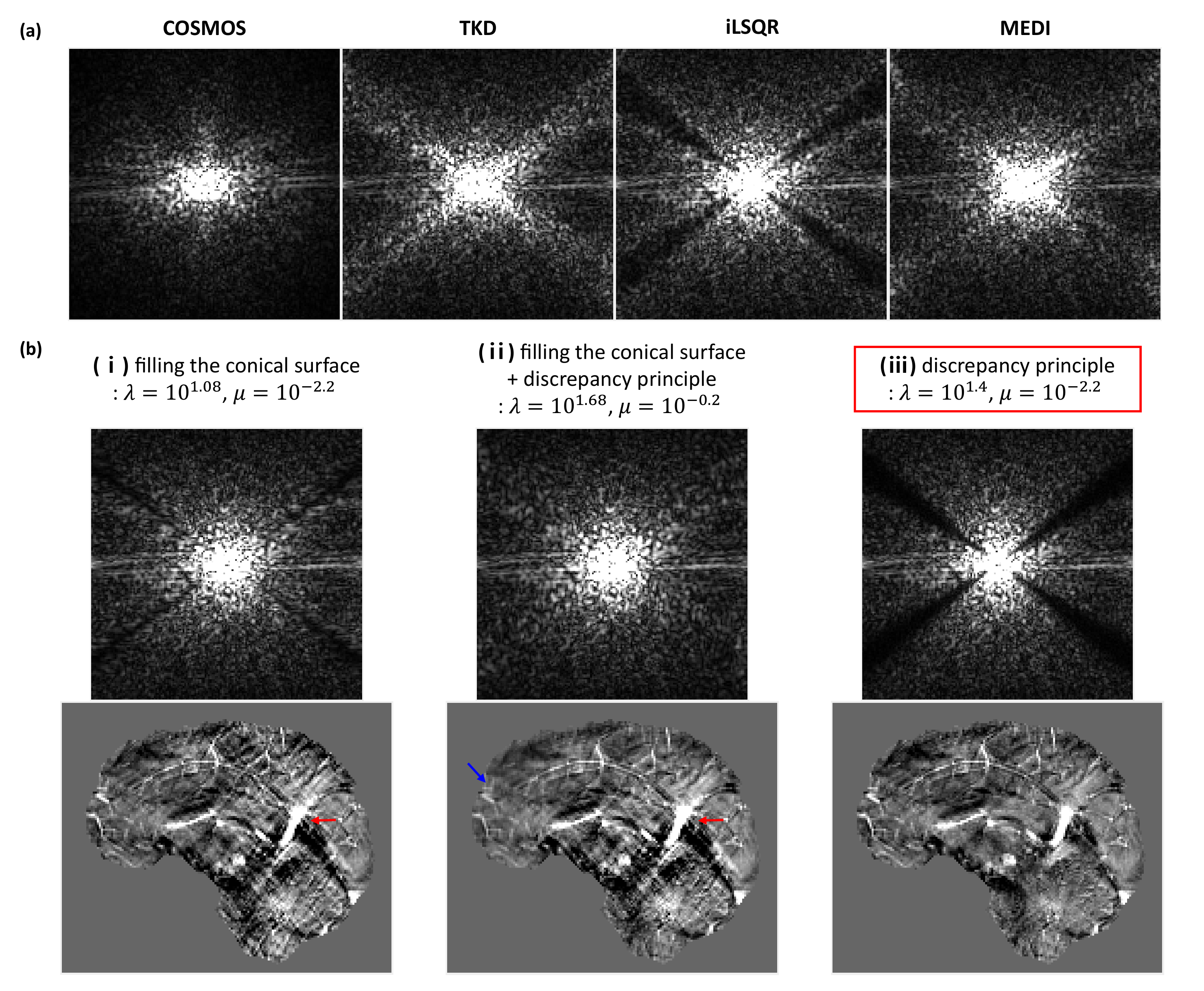}
\caption{
(a) Sagittal view of the reconstructed k-space with each QSM reconstruction method. (b) Sagtittal view of reconstructed ALOHA-QSM image and k-space. Parameter sets determined with three different criteria were compared.  
}\label{fig:ksp}
\end{figure*}

    Figure \ref{fig:ksp} shows a sagittal plane of a reconstructed k-space for each method. For ALOHA-QSM, signals were suppressed near the conical surface. The k-space of TKD and iLSQR also showed similar problems near the conical surface. 
	These unnatural energy level in k-space are caused by insufficient information on the conical surface of the dipole kernel filter, leading to systematic underestimation of the susceptibility values, which commonly happens in k-space based QSM reconstruction methods. Many studies were conducted to compensate for the underestimation. In TKD, the intensity of point spread function constructed from the difference between the original dipole kernel and the corresponding truncated kernel was used as correction factor \cite{F.Schweser.2013}. 
    In ALOHA-QSM, the correction factor was derived from the slope of linear regression between the original phase image and the phase image reformulated from the ALOHA-QSM reconstructed susceptibility image, compensating for the underestimation and thus providing more accurate susceptibility values.

The suppressed conical surface in ALOHA-QSM could be recovered by using smaller $\lambda$ for reconstruction. Based on this, we tested two additional criteria in comparison with the criteria of the discrepancy principle.
Results from the three criteria for determination of $\lambda$ and $\mu$ were compared in Fig.\ref{fig:ksp}b. 
For quantifying how much the k-space conical surface was filled, the pixels with top 5\% intensities were selected in the whole k-space and then the ratio of the top 5\% pixels belonging to the conical surface to those not belonging to the conical surface was measured as an indicator of the conical surface recovery. 
With $\lambda=10^{1.08}$, the conical surface was recovered well.
Also, the condition of $\lambda$ and $\mu$ satisfying the conical surface recovery and the discrepancy principle simultaneously ($\lambda=10^{1.68}, \mu=10^{-0.2}$) was compared with the two conditions satisfying them separately. 
The conditions filling the conical surface provided more blurry (blue arrow), artefactual structures (red arrow), and greater RMSE (2.69\%$\rightarrow$3.15,3.15\%) than the condition satisfying the discrepancy principle only. 
These results indicate that perfect recovery of signals near the conical surface does not necessarily mean reconstruction of the best QSM image in ALOHA-QSM. This is presumably due to noise boosting near the conical surface in this k-space based QSM reconstruction methods. 
The method referred to as consistency on the cone data (CCD) was suggested to enforce structural consistency on the conical surface with structural prior in the post-processing stage \cite{Y.Wen.2016}, and it recovered the conical surface in k-space well. We believe these methods complement each other and are synegistic.

\begin{figure*}[h!b]
\includegraphics[width=\textwidth]{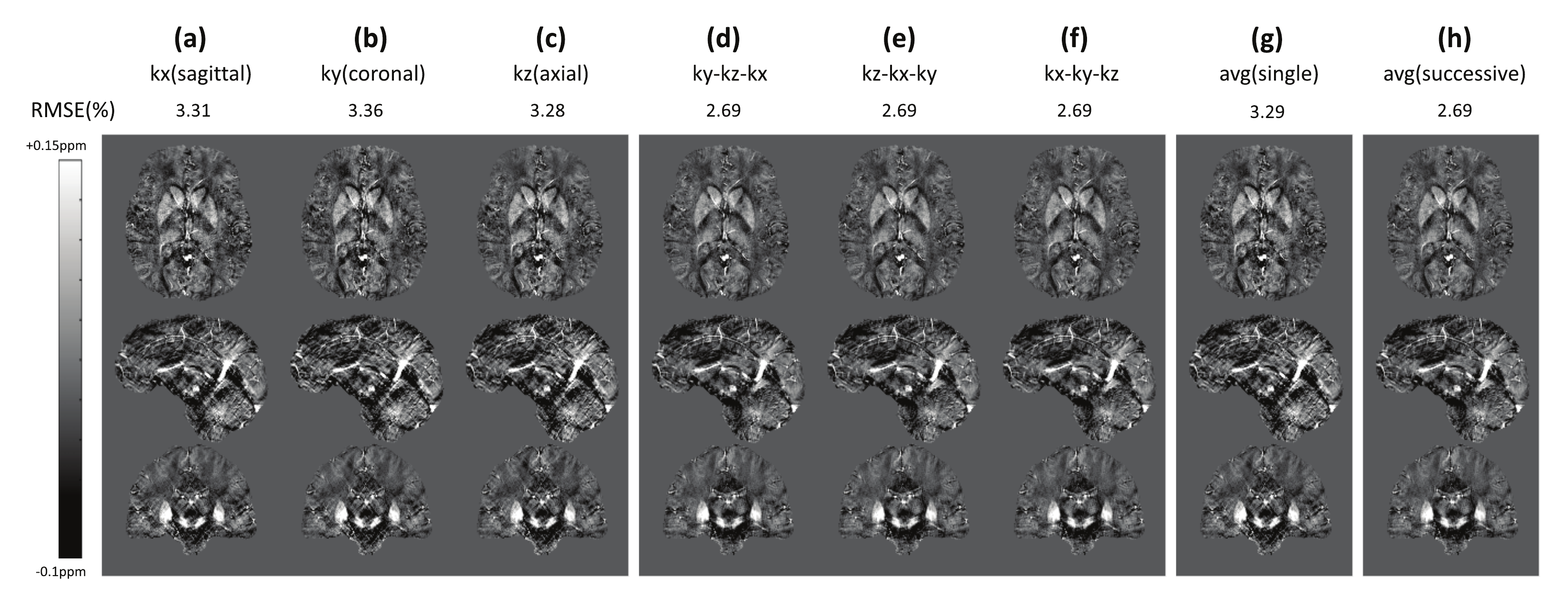}
\caption{
Comparison of \textit{in vivo} human brain susceptibility images reconstructed with different processing order of ALOHA-QSM. (a)-(c): reconstruction was implemented along single direction; (a) sagittal direction; (b) coronal direction; (c) axial direction. (d)-(f): ALOHA-QSM was implemented successively along (d) coronal-axial-sagittal direction; (e) axial-sagittal-coronal direction; (f) sagittal-coronal-axial direction. (g) Average of susceptibility images reconstructed with single direction. (h) Average of successively reconstructed results. 
}\label{fig:directional}
\end{figure*}

    The application of ALOHA has been extensively used for 2-D datasets due to the memory requirement of storing Hankel matrix. In this study, we resolved this issue by applying the ALOHA algorithm three times successively along three axes of the 3-D k-space. Successive implementation contributed to enhance processing speed and quality of the resulting image. When the algorithm was implemented along a single axis, images showed variations depending on the processing direction (Fig.\ref{fig:directional}a-c). This variation as well as RMSE decreased with successive reconstruction along the three directions (Fig.\ref{fig:directional}d-f). While the results were less dependent on the order of the three directions for successive reconstruction in our results, averaging all the results from the multiple orders can further enhance the data consistency and quality at the cost of increased processing time by a factor of three (Fig.\ref{fig:directional}g,h). As mentioned before, circulant approximation is recently proposed to handle large size 3-D Hankel matrix \cite{Ongie.2017(2)}. Application of this method may be an interesting research topic, which is beyond the scope of this study.


     Existing QSM reconstruction algorithms can be classified into two mainstreams: k-space based and magnitude image-guided methods. K-space based methods have a fast processing time and provide fine vascular structures, but are prone to streaking artifacts as shown for TKD in Fig.\ref{fig:comparison}. These streaking artifacts disturb evaluation of susceptibility images. 
	Magnitude image-guided methods such as MEDI use prior anatomical information for regularization and can generate susceptibility maps free of streaking artifacts. 
 However, structural boundaries on magnitude images are sometimes different from those on susceptibility maps, which may distort the results. For the simulated brain, MEDI showed better performance than any other methods, because structural boundaries on magnitude images exactly match those on susceptibility images including GM and WM which often show less clear boundaries (Fig.\ref{fig:comparison}a, Table.\ref{table:comparison}).
	From \textit{in vivo} data, details of brain cortex and small veins are over-smoothed and not depicted well in the susceptibility maps from MEDI (Fig.\ref{fig:comparison}b).
    Our suggested algorithm, ALOHA-QSM, reconstructed susceptibility maps with detailed structures, while suppressing streaking artifacts. ALOHA-QSM is derived by compressed sensing principle \cite{J.Ye.2017} and it can be classified into the k-space based method. 
	In this study, direct comparison of ALOHA-QSM against CS algorithms such as standard TV sparsity was not provided. A previous study showed that the TV approach distorts structural boundaries in images in a discernible way, while the distortions are significantly reduced in ALOHA \cite{K.Jin.2016}. 
	It can be explained in two reasons. The structured low-rank Hankel matrix approaches are based on the sampling theory of signals with finite rate of innovations on continuous domain. This is one of the important reasons to improve upon the TV approaches that are based on the discrete grid. Another important reason is that the structured low-rank Hankel matrix approach focus on recovering the edge signal, where TV approaches impose the piecewise continuity to the reconstructed images. Accordingly, ALOHA is different from the conventional CS approaches and is better in recovering edges, whereas edge distortions are usually observed in TV approaches.
    This new approach may open up a new direction on QSM reconstruction.

\section {Conclusion}    
    
    We demonstrated that the annihilating filter-based low-rank Hankel matrix constraint can be successfully used to solve the three-dimensional QSM dipole inversion problem. This new k-space based method, ALOHA-QSM, reconstructed accurate susceptibility maps with suppression of streaking and smoothing artifacts while requiring no anatomical constraint and exploiting its own phase information alone. 
    Therefore, it has high potential for QSM image reconstruction.

\section {Appendix}

\begin{algorithm}
\caption{Implementation Details on ALOHA-QSM}
\label{algorithm:ALOHAQSM}
\begin{algorithmic}
\STATE{Initialize with $\hat{\chi}=\hat{\varphi}/\hat{D}(\vec{k},a)$, where $\hat{D}(\vec{k},a)$ denotes dipole kernel truncated with threshold $a$.}
\FOR{$dir=[k_x,k_y,k_z]$}
    \FOR{$k=1,2,3,\cdots,n$}
        \IF{$dir=k_x$}
            \STATE $\hat{\chi}^{(k)}=\hat{\chi}(k,\vec{k_y},\vec{k_z})$
        \ELSIF{$dir=k_y$}
            \STATE
            $\hat{\chi}^{(k)}=\hat{\chi}(\vec{k_x},k,\vec{k_z})$
        \ELSIF{$dir=k_z$}
            \STATE
            $\hat{\chi}^{(k)}=\hat{\chi}(\vec{k_x},\vec{k_y},k)$
        \ENDIF
        \STATE
        $\hat{\chi}^{(k)}_w = \hat{\chi}^{(k)}\odot W_{Haar}$, where $W_{Haar}$ is the Fourier domain weighting filter of Haar wavelet transform.
        \STATE
        Solve $\min_{\hat{\chi'}^{(k)}_w}{\parallel \hat{\varphi}_w - \hat{D} \odot \hat{\chi'}^{(k)}_w \parallel}^2 + \lambda \textrm{RANK} (\mathcal{H}(\hat{\chi'}^{(k)}_w)$, 
        which initialized with $\hat{\chi}^{(k)}_w$.
        \STATE
        $\hat{\chi'}^{(k)}=\hat{\chi'}^{(k)}_w\odot W_{Haar}^{-1}$
        \STATE
        update $\hat{\chi}^{(k)}=\hat{\chi'}^{(k)}$
    \ENDFOR
\ENDFOR
\STATE{correction factor $s_m$ is the linear regression between $\varphi$ and $\varphi'$, where $\varphi'=FT^{-1}(\hat{D}(\vec{k})\odot\hat{\chi}(\vec{k}))$}
\STATE{Finally, $\chi=\chi/{s_m}$.}
\end{algorithmic}
\end{algorithm}

\section*{acknowledgements}
This work was supported by the National Research Foundation of Korea (NRF-2016R1A2B3008104, NRF-2017R1A2B2006526) and the Korea Health Technology R\&D Project through the Korea Health Industry Development Institute (KHIDI), funded by the Ministry of Health \& Welfare of South Korea (HI16C1111).

\bibliographystyle{IEEEtran}
\bibliography{main.bbl}
\newpage

\begin{table*}[!htb]
\centering
\caption{Regression coefficients ($R^2$, slope), root mean squared error (RMSE(\%)), and processing time for each reconstruction methods in simulated brain and \textit{in vivo} human brain. 
}
\label{table:comparison}
\begin{threeparttable}
\centering
\begin{tabular}{|l|cccc|cccc|}
\hline
\multicolumn{1}{|c|}{} & \multicolumn{4}{c|}{Simulated brain} & \multicolumn{4}{c|}{\textit{in vivo} Brain} \\ \hline
\multicolumn{1}{|c|}{} & $R^2$ & slope & RMSE(\%) & \begin{tabular}[c]{@{}c@{}}Processing\\ time (s)\end{tabular} & $R^2$ & slope & RMSE(\%) & \begin{tabular}[c]{@{}c@{}}Processing\\ time (s)\end{tabular} \\ \hline
TKD & 0.48 & 0.90 & 3.49 & 1.92 & 0.28 & 0.93 & 3.90 & 1.17 \\
iLSQR & 0.18 & 0.76 & 5.92 & 78.9 & 0.40 & 1.02 & 3.31 & 68.3 \\
MEDI & 0.72 & 0.81 & 1.96 & 470 & 0.30 & 0.95 & 3.85 & 302 \\
\begin{tabular}[c]{@{}l@{}}ALOHA-QSM$^{*}$\\ (no correct)\end{tabular} & \multirow{2}{*}{0.62} & 0.56 & 2.39 & \multirow{2}{*}{187} & \multirow{2}{*}{0.43} & 0.62 & 2.12 & \multirow{2}{*}{156} \\
\begin{tabular}[c]{@{}l@{}}ALOHA-QSM$^{*}$\\ (corrected)\end{tabular} &  & 0.77 & 2.46 &  &  & 0.89 & 2.69 &  \\ \hline
\end{tabular}
\begin{tablenotes}
\item{*} GPU was used to accelerate algorithm.
\end{tablenotes}
\end{threeparttable}
\end{table*}

\begin{table*}[ht]
\centering
\caption{Comparison of the estimated mean susceptibility (ppm) and standard deviation for deep gray matter structures with literature.}
\label{table:roi}
\begin{adjustbox}{width=\textwidth}
\begin{tabular}{l|c|ccccc}
\hline
\multicolumn{1}{c|}{\textbf{Method}} & \textbf{N} & \textbf{Substantia nigra} & \textbf{Red nucleus} & \textbf{Globus pallidus} & \textbf{Putamen} & \textbf{Caudate nucleus} \\ \hline
ALOHA-QSM (current study) & 5 & 0.10$\pm$0.04 & 0.09$\pm$0.01 & 0.12$\pm$0.02 & 0.05$\pm$0.01 & 0.05$\pm$0.01 \\
COSMOS \cite{Lim.2013} & 5 & 0.09$\pm$0.01 & 0.06$\pm$0.00 & 0.11$\pm$0.01 & 0.03$\pm$0.01 & 0.02$\pm$0.00 \\
COSMOS \cite{T.Liu.2011} & 9 & 0.13$\pm$0.03 & 0.09$\pm$0.02 & 0.19$\pm$0.02 & 0.09$\pm$0.04 & 0.08$\pm$0.02 \\
\begin{tabular}[c]{@{}l@{}}COSMOS \\ \cite{S.Wharton.2010(1)}\end{tabular} & 5 & 0.17$\pm$0.02 & 0.14$\pm$0.02 & 0.19$\pm$0.02 & 0.10$\pm$0.01 & 0.09$\pm$0.02 \\
L2 RSO \cite{B.Bilgic.2012} & 11 & 0.08$\pm$0.04 & 0.07$\pm$0.03 & 0.12$\pm$0.02 & 0.06$\pm$0.02 & 0.07$\pm$0.02 \\
\begin{tabular}[c]{@{}l@{}}L2 RSO\\ \cite{S.Wharton.2010(1)}\end{tabular} & 5 & 0.16$\pm$0.03 & 0.13$\pm$0.02 & 0.19$\pm$0.02 & 0.09$\pm$0.01 & 0.09$\pm$0.01 \\
MEDI \cite{B.Bilgic.2012} & 11 & 0.10$\pm$0.04 & 0.09$\pm$0.04 & 0.14$\pm$0.02 & 0.07$\pm$0.02 & 0.08$\pm$0.02 \\
MEDI \cite{T.Liu.2011} & 9 & 0.12$\pm$0.03 & 0.08$\pm$0.05 & 0.19$\pm$0.02 & 0.08$\pm$0.02 & 0.09$\pm$0.02 \\ \hline
\end{tabular}
\end{adjustbox}
\end{table*}

\end{document}